\documentclass[letterpaper,10pt,conference]{ieeeconf} % Comment this line out if you need a4paper

\IEEEoverridecommandlockouts % This command is only needed if
\usepackage{amsmath,amsfonts,amssymb}
\usepackage{graphicx}
\usepackage[top=1in, bottom=1in, left=1in, right=1in]{geometry}
\usepackage{bm}

\usepackage{hyperref}
\hypersetup{
pdfstartview={FitH}, breaklinks=true,
colorlinks=true, linkcolor=black,
citecolor=black, filecolor=black,
urlcolor=black,
}

\title{\LARGE \bf Markov Data-Based Reference Tracking of Tensegrity Morphing Airfoils}
\author{Yuling Shen$^{1}$, Muhao Chen$^{1}$, Maoranjan Majji$^{2}$, and Robert E. Skelton$^{2}$% <-this % stops a space
\thanks{*This work was not supported by any organization.}% <-this % stops a space
\thanks{$^{1}$ Yuling Shen and Muhao Chen are graduate students with the Department of Aerospace Engineering, Texas A\&M University, College Station, USA.
{\tt\small yus011@tamu.edu, muhaochen@tamu.edu.}}%
\thanks{$^{2}$ Manoranjan Majji is the director of LASR Laboratory, assistant professor and Robert E. Skelton is a TEES eminent professor with the Department of Aerospace Engineering, Texas A\&M University, College Station, USA.
{\tt\small mmajji@tamu.edu, bobskelton@tamu.edu.}}}

\begin{document}

\maketitle
\thispagestyle{empty}
\pagestyle{empty}

%%%%%%%%%%%%%%%%%%%%%%%%%%%%%%%%%%%%%%%%%%%%%%%%%%%%%%%%%%%%%%%%%%%%%%%%%%%%%%%%
\begin{abstract}
This letter presents a data-based control design for reference tracking applications. This design finds the optimal control sequence, which minimizes a quadratic cost function consisting of tracking error and input increments over a finite interval $[0,N]$. The only information needed is the first $N+1$ Markov parameters of the system. This design is employed on a tensegrity morphing airfoil whose topology has been described in detail in this letter. A NACA 2412 airfoil with specified morphing targets is chosen to verify the developed design. The principle developed in this letter is also applicable to other structural control problems.
\end{abstract}
%%%%%%%%%%%%%%%%%%%%%%%%%%%%%%%%%%%%%%%%%%%%%%%%%%%%%%%%%%%%%%%%%%%%%%%%%%%%%%%%
\section{Introduction}

Researchers used to formulate the control problems of dynamic systems by starting with their equations of motions. For a nonlinear system, an appropriate linearization can be found, and most of the linear control theories become applicable. However, we simply are not able to write down the dynamics of systems of interest every time (such as black box systems) or do not trust the dynamics that we have already had. The development of modern technology enables the vast storage and fast process of big data, thereby data-driven approaches relying on input-output data have emerged. Lim and Phan developed an observer from I/O data, which estimates the state of the system at some future step \cite{lim1997identification}. Safonov and Taso developed a method to determine a validated control law that meets given performance specifications from I/O data \cite{safonov1997unfalsified}. Zhang et al. developed a data-driven control approach that recognizes a neural network model from I/O data then apply adaptive dynamic programming on it \cite{zhang2011data}. Proctor et al. used the technique of regression that recognizes a model from I/O data \cite{proctor2016dynamic}. However, most of these approaches seek the best fit for input-output data, which may have no explicit physical explanations. On the other side, a few attempts have been made to control a system with interpreted physical properties from I/O data, such as Markov parameters. Markov parameters, which are impulse responses of a system, are evaluated from the knowledge of input-output data, step response \cite{ljung1987system}, or well-conditioned time response \cite{juang1994applied}. The work by Furuta et al. solves the finite horizon LQG problem using an infinite number of Markov parameters \cite{furuta1993dynamic}. Built upon that,  Shi and Skelton proposed their Markov data-based control design in 2000 \cite{shi2000markov}, which reduces the required information to the first $N+1$ Markov parameters only. While these designs are limited to regulator applications, this letter extends the work of Shi and Skelton by presenting a new Markov data-based control design for reference tracking applications, with a requirement of the first $N+1$ Markov parameters of a system. \par

%It is also inevitable that the dynamics of a physical system will change as time goes by due to reasons such as material fatigue, erosion, friction, or any other internal reactions that are unforeseen.

%While most of these approaches seek a least-square solution, an optimal control law based on the most updated properties of a physical system without the reliability of the system dynamics would be useful. 

% The Markov data-based control design, proposed by, is derived from a model-based LQG control law. It optimizes a control sequence that minimizes a quadratic cost function of system output and control input over a finite horizon. When extending to reference tracking applications, a new cost function shall be designed. In most cases, at least some input is required to hold the system at a reference target, which is not an equilibrium point. The model-based control has been modified through re-assemble the system by introducing the input increment as the new input signal \cite{rawlings2017model}. The new model-based control law minimizes tracking error with steady control input, and so does its data-based derivation.

Tensegrity structure is prestressable and stable structures composed of compressive (bars/struts) and tensile members (strings/cables) \cite{skelton2009tensegrity}. Biological systems provide perhaps the greatest evidence that tensegrity concepts yield the most efficient structures. For example, the molecular structure of cell surface \cite{wang2001mechanical}, DNA bundles \cite{liedl2010self}, spider fibers \cite{simmons1996molecular}, and human elbows \cite{scarr2012consideration} are internally consistent with tensegrity models. After decades of study, tensegrity systems has shown its advantages in properties such as lightweight \cite{chen2020general,ma2020design}, deployability \cite{chen2020deployable,yildiz2020deployment}, energy absorption \cite{garanger2020soft,sunny2014optimal}, and promoting integration of structural and control \cite{chen2020habitat,li2020new,sabelhaus2020inverse}, etc. 

Due to the many benefits of tensegrity systems, many soft robots have been developed by control and robotics communities using the tensegrity paradigm in recent years. For example, Baines et al. presented tensegrity rolling robots by soft membrane actuators \cite{baines2020rolling}. Wang et al. derived a nonlinear dynamics-based control and a decoupled data-based (D2C) LQR control law around the linearized open-loop trajectories \cite{wang2020model}. Goyal et al. developed tensegrity robotics with gyros as the actuators \cite{goyal2020gyroscopic}. Begey et al. demonstrated two approaches, X-shaped tensegrity mechanisms and an analogy with scissor structures, for the design of tensegrity-based manipulators \cite{begey2020design}.

Comparing with rigid airfoils, morphing airfoils are gaining significant interest from various researchers due to its promising advantages in flexibility to different flight regimes. Although current approaches, such as flaps, slats, aileron, and wing-let, can help achieve the desired control objective, most of these efforts are basically breaking the streamline airfoil shape, which is carefully designed by aerodynamics engineers. Other emerging methods, such as shape memory alloys, piezoelectric actuators, can also fulfill the morphing targets smoothly. However, most of them working at relatively low bandwidth and require heavy supporting equipment. A few attempts have also been made towards a system point of view by integrating structure and control designs. For example, Chen et al. \cite{chen2020design} presented a design of tensegrity morphing airfoil and nonlinear control approach to class-$k$ tensegrity structures. Shintake et al. designed a fish-like robot with tensegrity systems, which are driven by a waterproof servomotor \cite{shintake2020bio}. This letter also tries to demonstrate structural control with an application of a tensegrity morphing airfoil control by a data-based approach.

The letter is structured as follows: Section \ref{sec2} introduces the Markov data-based control design. Section \ref{sec3} describes the topology of the tensegrity morphing airfoil. Section \ref{sec4} provides an explanation of the morphing target. Section \ref{sec5} shows the results developed from the control design. Finally, a conclusion and a brief discussion about the letter is given in section \ref{sec6}.

% Introduction:
% \begin{itemize}
%     \item Model based control and Data-Based control theories
%     \item what is data-based control
%     \item why we need data-based control law 
%     \item what are the categories, difference, PID, machine learning, ect. 
%     \item what is tensegrity, why 
%     \item contribution of this letter
% \end{itemize}

% Here is the scope of this letter. 
% \begin{itemize}
%     \item Introduction
%     \item Data-based control law
%     \item Tensegrity airfoil 
%     \item Morphing target
%     \item Data experiment and results
%     \item Conclusion
% \end{itemize}

% In 1903, the fundamental breakthrough of the Wright Brothers' 

% A misconception is that “The best system is made from the best components.” That’s certainly not true. Often, we gain more in integrating two disciplines than making exceptional improvements in one discipline. For example, the airplane wing. Aerodynamics guy first designed the best shape based on their knowledge in fluid dynamics, then the control guy came and break the beautiful shape for control objectives. This is certainly not the right way. A systematic approach would be to Modify shape not by pushing against a reference equilibrium but by modifying the equilibrium! Of course, this would require much less control effort. 

%%%%%%%%%%%%%%%%%%%%%%%%%%%%%%%%%%%%%%%%%%%%%%%%%%%%%%%%%%%%%%%%%%%%%%%%%%%%%%%%%%%%%%%%%%%%%%%%%%%%%%%%%%%%%%%%%%%%%%%%%%%%%%%%%%

\section{Data-Based Control Law}\label{sec2}

This section first derives a model-based optimal control design for reference tracking applications that minimizes a quadratic cost function consists of tracking error and control increment. It is transformed into a data-based design which requires no additional information about the dynamics of the system but the first $N+1$ Markov parameters.\par

\subsection{Model Based Optimal Control}
\label{section_model_based}

Consider the following state space system:
\begin{equation}
\label{oldsystem}
    G:\left\{
    \begin{array}{rl}
    x_{k+1} &= Ax_k+B(u_k+w_k)\\
    y_k &= Cx_k\\
     \end{array}
    \right.,
\end{equation}
where $A$, $B$, $C$ are state space coefficient matrices, $u_k$ is the control input, $w_k$ is the input disturbance and $v_k$ is the sensor noise. \par
A regulator application finds an optimal control sequence which minimizes input and output with some weights. In reference tracking applications, this does not apply since holding system to a reference configuration may require some actuation. One solution would be to modify the cost function. Consider a new system as the following \cite{rawlings2017model}:
\begin{equation}
\label{newsystem}
    \hat{G}:\left\{
     \begin{array}{rl}
     \hat{x}_{k+1} &=\hat{A} \hat{x}_{k}+\hat{B}\Delta u_k+\hat{D}w_k,\\
     y_{k} &=\hat{C}\hat{x}_{k}+v_{k},\\
    \end{array}
    \right.,
\end{equation}
where $\Delta u_{k} = u_k - u_{k-1}$ is the input increment, and
\begin{align}
    \hat{A} & = \begin{bmatrix}A & B \\
0 & I\end{bmatrix}, \hat{B}=\begin{bmatrix}B \\
I\end{bmatrix},\hat{D}=\begin{bmatrix}D \\
0\end{bmatrix}, \\
\hat{C} & =\begin{bmatrix}C & 0\end{bmatrix},\hat{x}_{k}=\begin{bmatrix}x_{k} \\ u_{k-1}\end{bmatrix}.
\end{align}

Given a trajectory reference signal $r_k$ which the output $y_k$ matches to. A cost function may be selected as the following:
\begin{align}
\nonumber
J   = & \frac{1}{2}(r_{N}-y_{N})^{T} S(r_{N}-y_{N})   +\frac{1}{2} \sum_{k=0}^{N-1}[(r_{k}-y_{k})^{T} Q(r_{k}-y_{k}) \\ \label{costfunction}
 & +\Delta u_{k}^{T} R \Delta u_{k}]   + \frac{1}{2}\Delta u_{N}^{T} T\Delta u_{N}^{T},
\end{align}
which minimizes the accumulation of tracking error and and rate of change of control input. The solution can be computed by taking $\nabla J = 0$. The solution of input increment sequence is given as below \cite{rawlings2017model}: 
\begin{align}
    \begin{bmatrix} \Delta u_0\\ \Delta u_1\\ \vdots \\  \Delta u_N\end{bmatrix} & = \mathbf{K}_0\left(\begin{bmatrix}  r_0\\  r_1\\ \vdots \\   r_N\end{bmatrix} - \begin{bmatrix} \hat{C}\\ \hat{C}\hat{A}\\ \vdots \\  \hat{C}\hat{A}^N\end{bmatrix} \hat{x}_0\right), \\
 \mathbf{K}_0 & = (\mathbf{\bar{H}}_0^T\mathbf{\bar{Q}}_0\mathbf{\bar{H}}_0+\mathbf{\bar{R}}_0)^{-1}(\mathbf{\bar{Q}}_0\mathbf{\bar{H}}_0)^T,\\
 \mathbf{\bar{H}}_0 & = \begin{bmatrix}0 & 0 & \cdots & \cdots & 0\\\hat{C}\hat{B} & 0 & \ddots & \ddots & 0 \\\hat{C}\hat{A}\hat{B} & \hat{C}\hat{B} & \ddots & \ddots & \vdots\\\vdots & \vdots & \ddots & 0 & \vdots \\ \hat{C}\hat{A}^{N-1}\hat{B} & \hat{C}\hat{A}^{N-2}\hat{B} & \cdots & \hat{C}\hat{B} & 0 \end{bmatrix}, \\
 \mathbf{\bar{Q}}_0  & = \begin{bmatrix}Q & 0 & \cdots & 0\\ 0 & \ddots & \ddots & \vdots \\\vdots & \ddots & Q & 0\\0 & \cdots & 0 & S\end{bmatrix},~\mathbf{\bar{R}}_0 = \begin{bmatrix}R & 0 & \cdots & 0\\ 0 & \ddots & \ddots & \vdots \\\vdots & \ddots & R & 0\\0 & \cdots & 0 & T\end{bmatrix},
\end{align}
where the subscript $0$ represents the step $0$ within a horizon of $N$. The notation of bold font ($\mathbf{\bar{H}}_0$, $\mathbf{\bar{Q}}_0$, and $\mathbf{\bar{R}}_0$) indicates that these parameters will change according to their index number. Now consider a more general formula for the step $k \in [1, N]$. Without loss of generality, the following equation can be established \cite{shi2000markov}:
\begin{align}
\label{control_law_uk}
    \begin{bmatrix} \Delta u_k\\ \Delta u_{k+1}\\ \vdots \\  \Delta u_N\end{bmatrix} & = \mathbf{K}_k \left(\begin{bmatrix}  r_k\\  r_{k+1}\\ \vdots \\   r_N\end{bmatrix} - \begin{bmatrix} \hat{C}\\ \hat{C}\hat{A}\\ \vdots \\  \hat{C}\hat{A}^{N-k}\end{bmatrix}\hat{x}_k\right), \\
\mathbf{K}_k &= (\mathbf{\bar{H}}_k^T\mathbf{\bar{Q}}_k\mathbf{\bar{H}}_k+\mathbf{\bar{R}}_k)^{-1}(\mathbf{\bar{Q}}_k\mathbf{\bar{H}}_k)^T, 
\end{align}
where $\mathbf{\bar{H}}_k$, $\mathbf{\bar{Q}}_k$, and $\mathbf{\bar{R}}_k$ are given as:
\begin{align}
\mathbf{\bar{H}}_k & = \begin{bmatrix}0 & 0 & \cdots & \cdots & 0\\\hat{C}\hat{B} & 0 & \ddots & \ddots &  0\\\hat{C}\hat{A}\hat{B} & \hat{C}\hat{B} & \ddots & \ddots & \vdots\\\vdots & \vdots & \ddots & 0 & 0 \\ \hat{C}\hat{A}^{N-k-1}\hat{B} & \hat{C}\hat{A}^{N-k-2}\hat{B} & \cdots & \hat{C}\hat{B} & 0 \end{bmatrix},\\ \label{weights}
\mathbf{\bar{Q}}_k =& \begin{bmatrix}Q & 0 & \cdots & 0\\ 0 & \ddots & \ddots & \vdots \\\vdots & \ddots & Q & 0\\0 & \cdots & 0 & S\end{bmatrix}, \mathbf{\bar{R}}_k = \begin{bmatrix}R & 0 & \cdots & 0\\ 0 & \ddots & \ddots & \vdots \\\vdots & \ddots & R & 0 \\0 & \cdots & 0 & T\end{bmatrix}.
\end{align}

\subsection{State Estimator}
\label{section_state}
The parameter $\mathbf{H}_k$ can be formulated using Markov parameters $\hat{H}_i$ (see section \ref{section_markov}). The remaining unknown part is defined as the following variable:
\begin{align}
    \bar{x}_k & =\begin{bmatrix} \hat{C}\\ \hat{C}\hat{A}\\ \vdots \\  \hat{C}\hat{A}^{N-k}\end{bmatrix}\check{x}_k,\\
    \check{x}_{k+1} & = \hat{A}\check{x}_k + \hat{B}\Delta u_k + \mathbf{L}_k(y_k-\hat{C}\check{x}_k),
\end{align}
where $\check{x}_k$ is the estimation of the state $\hat{x}_k$, and $L_k$ stands for the gain of the estimator. The following relation can be established \cite{shi2000markov}:
\begin{align}\nonumber
\bar{x}_{k} & =  \begin{bmatrix} \hat{C}\\ \hat{C}\hat{A}\\ \vdots \\  \hat{C}\hat{A}^{N-k+1}\end{bmatrix}\check{x}_{k-1}  +\begin{bmatrix} \hat{C}\hat{B}\\\hat{C}\hat{A}\hat{B}\\ \vdots \\\hat{C}\hat{A}^{N-k\hat{B}}\end{bmatrix}\Delta u_{k-1} \\ & +\begin{bmatrix} \hat{C}\\ \hat{C}\hat{A}\\ \vdots \\  \hat{C}\hat{A}^{N-k}\end{bmatrix}\mathbf{L}_{k-1}(y_{k-1}-\hat{C}\check{x}_{k-1}) \\
 &= \nonumber\begin{bmatrix} -\begin{bmatrix} \hat{C}\\ \hat{C}\hat{A}\\ \vdots \\  \hat{C}\hat{A}^{N-k}\end{bmatrix}\mathbf{L}_{k-1} & I\end{bmatrix}\begin{bmatrix} \hat{C}\\ \hat{C}\hat{A}\\ \vdots \\  \hat{C}\hat{A}^{N-(k-1)}\end{bmatrix} \check{x}_{k-1}\\& +v_{k} + \begin{bmatrix} \hat{C}\hat{B}\\\hat{C}\hat{A}\hat{B}\\ \vdots \\\hat{C}\hat{A}^{N-k}\hat{B}\end{bmatrix}\Delta u_{k-1} + \begin{bmatrix} \hat{C}\\ \hat{C}\hat{A}\\ \vdots \\  \hat{C}\hat{A}^{N-k}\end{bmatrix}\mathbf{L}_{k-1}y_{k-1}\\
 &= \begin{bmatrix}-\mathbf{F}_k & I
 \end{bmatrix}\bar{x}_{k-1} + \mathbf{B}_k\Delta u_{k-1} + \mathbf{F}_k y_{k-1},
\end{align}
where
\begin{align}
  \mathbf{F}_k = \begin{bmatrix} \hat{C}\\ \hat{C}\hat{A}\\ \vdots \\  \hat{C}\hat{A}^{N-k}\end{bmatrix}\mathbf{L}_{k-1},\quad \mathbf{B}_k=\begin{bmatrix} \hat{C}\hat{B}\\\hat{C}\hat{A}\hat{B}\\ \vdots \\\hat{C}\hat{A}^{N-k}\hat{B}\end{bmatrix}.  
\end{align}

The solution of the estimator gain can be expressed as the following \cite{furuta1993closed}:
\begin{align}
\label{estimator}
    \mathbf{L}_{k-1} &= \hat{A}\mathbf{Y}_{k-1}\hat{C}^T(V+\hat{C}\mathbf{Y}_{k-1}\hat{C}^T)^{-1},\\\nonumber
        \mathbf{Y}_{k-1} &= \mathbf{D}_{k-1}\mathbf{P}_k\mathbf{D}_{k-1}^T,\\
\mathbf{D}_{k-1} &= \begin{bmatrix}\hat{D} & \hat{A}\hat{D} & \cdots & \hat{A}^{k-1}\hat{D}
\end{bmatrix}, \\
\mathbf{P}_k &= (\mathbf{W}_{k-1}^{-1} + \mathbf{T}_{k-1}^T\mathbf{V}_{k-1}^{-1}\mathbf{T}_{k-1})^{-1},\\
\mathbf{W}_{k-1}&=diag(W),\quad W = E(w_k^Tw_k),\\
\mathbf{V}_{k-1}&=diag(V),\quad V = E(v_k^Tv_k),\\
\mathbf{T}_{k-1} &= \begin{bmatrix}0 & \hat{C}\hat{D} & \cdots & \cdots & \hat{C}\hat{A}^{k-2}\hat{D}\\0 & 0 & \ddots & \ddots & \hat{C}\hat{A}^{k-3}\hat{D}\\\vdots & \ddots & \ddots & \ddots & \vdots\\\vdots & \ddots & \ddots & \ddots & \hat{C}\hat{D} \\ 0 & 0 & \cdots & 0 & 0 \end{bmatrix}.
\end{align}

Applying Eq. (\ref{estimator}) one may get the following expression:
\begin{equation}
    \mathbf{F}_k=\mathbf{M}_k\mathbf{P}_k\mathbf{N}_k^T(V+\mathbf{N}_k\mathbf{P}_k\mathbf{N}_k^T)^{-1},
\end{equation}
\begin{align}\nonumber
    \mathbf{M}_k&=\begin{bmatrix}\hat{C}\hat{A}\hat{D} & \hat{C}\hat{A}^2\hat{D} & \dots & \hat{C}\hat{A}^k\hat{D}\\\hat{C}\hat{A}^2\hat{D} & \hat{C}\hat{A}^3\hat{D} & \dots & \hat{C}\hat{A}^{k+1}\hat{D}\\\vdots & \vdots & \ddots & \vdots\\\hat{C}\hat{A}^{N-k+1}\hat{D} & \hat{C}\hat{A}^{N-k+2}\hat{D} & \cdots & \hat{C}\hat{A}^N\hat{D} \end{bmatrix},\\
    \mathbf{N}_k &= \begin{bmatrix}\hat{C}\hat{D} & \hat{C}\hat{A}\hat{D} & \cdots & \hat{C}\hat{A}^{k-1}\hat{D}
\end{bmatrix}.
\end{align}\par
With the implementation of section \ref{section_markov}, all parameters can be formulated using Markov parameters of $\hat{H}_i$ and $\hat{M}_i$.

\subsection{Markov Parameters}
\label{section_markov}

Markov parameters of a system convey its transient properties and can be evaluated via many approaches. A common approach is to find the impulse responses of a system which is identical to Markov parameters. Besides that, two approaches are introduced below.\par

Given an unknown system, one may determine its $i$th Markov parameter regarding input ($H_i$) and disturbance ($M_i$) experimentally from the following input-output relation:
\begin{align}\label{H_matrix}
    H_i & = E(y_{k+i}u_k^T) = \lim_{N\to\infty} \sum_{k=0}^{N-1} y_{k+i}u_k^T,\\
    M_i & = H_i,
\end{align}
where $u_k$ is a white noise input signal with covariance $U=I$, and $y_k$ is the output signal. For a linear system $G$ which we know the exact dynamics, its Markov parameters may be computed via state space coefficients as the following:
\begin{equation}
    H_i = M_i = \left\{
\begin{array}{cc}
   0 &,~i=0\\
   CA^{i-1}B &,~i>0\\
   \end{array}
   \right..
\end{equation}

Similarly, the Markov parameters for the augmented system $\hat{G}$ are the following:
\begin{align}
    \hat{H}_i & = \left\{
\begin{array}{cc}
 0 &,~i=0\\
\hat{C}\hat{A}^{i-1}\hat{B} &,~i>0\\
\end{array}
 \right.,\\
\hat{M}_i & = \left\{
\begin{array}{cc}
0 &,~i=0\\
\hat{C}\hat{A}^{i-1}\hat{D} &,~i>0\\
\end{array}
 \right..
\end{align}

The following pattern can be observed:
\begin{align}
    \hat{H_i} & = \sum_{j=0}^{i} H_j,~ \hat{M_i} = M_i.
\end{align}

Therefore, one may get Markov parameters of the augmented system $\hat{G}$ from that of $G$, which can be determined experimentally. It is also obvious that parameters of $\mathbf{\bar{H}}_k$,  $\mathbf{B}_{k}$, $\mathbf{M}_{k}$, and $\mathbf{N}_{k}$ can be constructed if one knows the Markov sequences $\{H_0,H_1,..., H_{N-1}\}$ and $\{M_0,M_1,...,M_{N+1}\}$.

\subsection{Data-Based Control Law}

Combining the results from sections \ref{section_model_based}, \ref{section_state} and \ref{section_markov}, the complete Markov data-based reference tracking design for a system $G$ is summarized here. Notice the only required information is the Markov sequences $\{H_0,H_1,..., H_{N-1}\}$ and $\{M_0,M_1,...,M_{N+1}\}$.
\begin{equation}
u_k = \left\{
    \begin{array}{cc}
    0 &, k=0\\
    u_{k-1} + \Delta u_k &~~~~~, k\in[1,N]\\ 
    \end{array}
    \right.,
\end{equation}
where we have:
\begin{align}\label{delta_control}
\begin{bmatrix} \Delta u_k\\ \Delta u_{k+1}\\ \vdots \\  \Delta u_N\end{bmatrix} &  = \mathbf{K}_k \left(\begin{bmatrix}  r_k\\  r_{k+1}\\ \vdots \\   r_N\end{bmatrix} - \bar{x}_k\right),\\
\mathbf{K}_k & = (\mathbf{\bar{H}}_k^T\mathbf{\bar{Q}}_k\mathbf{\bar{H}}_k+\mathbf{\bar{R}}_k)^{-1}(\mathbf{\bar{Q}}_k\mathbf{\bar{H}}_k)^T.
\end{align}
The parameters $\mathbf{\bar{Q}}_k$, and $\mathbf{\bar{R}}_k$ are given in Eq. (\ref{weights}). $\mathbf{\bar{H}}_k$ is given below:
\begin{align}
\mathbf{\bar{H}}_k & = \begin{bmatrix}\hat{H}_0 & 0 & \cdots & \cdots & 0\\\hat{H}_1 & \hat{H}_0 & \ddots & \ddots & \vdots\\\hat{H}_2 & \hat{H}_1 & \ddots & \ddots & \vdots\\\vdots & \vdots & \ddots & \ddots & 0 \\ \hat{H}_{N-k} & \hat{H}_{N-k-1} & \dots & \hat{H}_1 & \hat{H}_0 \end{bmatrix}.
\end{align}
The rule of update for $\bar{x}_{k}$ is:
\begin{align} 
    \bar{x}_{k} & = \left\{
    \begin{array}{cc}
        0 &, k=0\\
     \begin{bmatrix}-\mathbf{F}_k & I\end{bmatrix}\bar{x}_{k-1} + \mathbf{B}_k\Delta u_{k-1} + \mathbf{F}_ky_{k-1} &~~~~~, k\in[1,N]\\ 
    \end{array}
    \right.,\\
\mathbf{F}_k &  = \mathbf{M}_k\mathbf{P}_k\mathbf{N}_k^T(V+\mathbf{N}_k\mathbf{P}_k\mathbf{N}_k^T)^{-1},\end{align}
\begin{align} 
\mathbf{B}_k &  = \begin{bmatrix}\hat{H}_1 \\ \hat{H}_2 \\\vdots\\\hat{H}_{N-k+1}
\end{bmatrix}, \\
    \mathbf{M}_k & = \begin{bmatrix}\hat{M}_2 & \hat{M}_3 & \cdots & \hat{M}_{k+1}\\\hat{M}_3 & \hat{M}_4 & \cdots & \hat{M}_{k+2}\\\vdots & \vdots & \ddots & \vdots\\\hat{M}_{N-k+2} & \hat{M}_{N-k+3} & \dots & \hat{M}_{N+1} \end{bmatrix}, \\
\mathbf{N}_k & = \begin{bmatrix}\hat{M}_1 & \hat{M}_2 & \cdots & \hat{M}_k
\end{bmatrix}, \\
\mathbf{P}_k &= (\mathbf{W}_{k-1}^{-1} + \mathbf{T}_{k-1}^T\mathbf{V}_{k-1}^{-1}\mathbf{T}_{k-1})^{-1},\\
\mathbf{W}_{k-1}&=diag(W),\quad W = E(w_k^Tw_k),\\
\mathbf{V}_{k-1}&=diag(V),\quad V = E(v_k^Tv_k),\\
\mathbf{T}_{k-1} &= \begin{bmatrix}\hat{M_0} & \hat{M_1} & \cdots & \cdots & \hat{M}_{k-1}\\ 0 & \hat{M_0} & \ddots & \ddots & \hat{M}_{k-2}\\\vdots & \ddots & \ddots & \ddots & \vdots\\\vdots & \ddots & \ddots & \ddots & \hat{M_1} \\ 0 & \cdots & \cdots & 0 & \hat{M_0} \end{bmatrix}.\\
\hat{H_i} & = \sum_{j=0}^{i} H_j,~
\hat{M_i} = M_i.
\end{align}

\section{Tensegrity Morphing Airfoil}\label{sec3}

\begin{figure}[htb]
\centering
\includegraphics[width=1.0\linewidth]{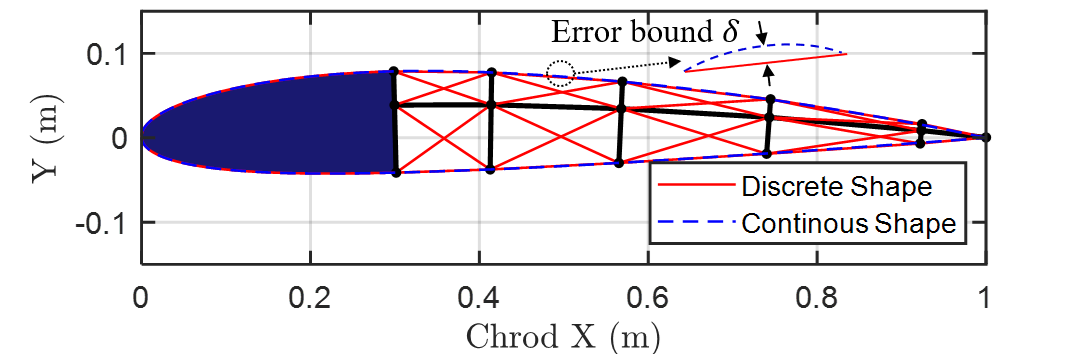}
\caption{Tensegrity airfoil configuration, blue area is the rigid body, black and red lines are bars and strings.}
\label{Config}
\end{figure}

\begin{figure}[htb]
\centering
\includegraphics[width=1.0\linewidth]{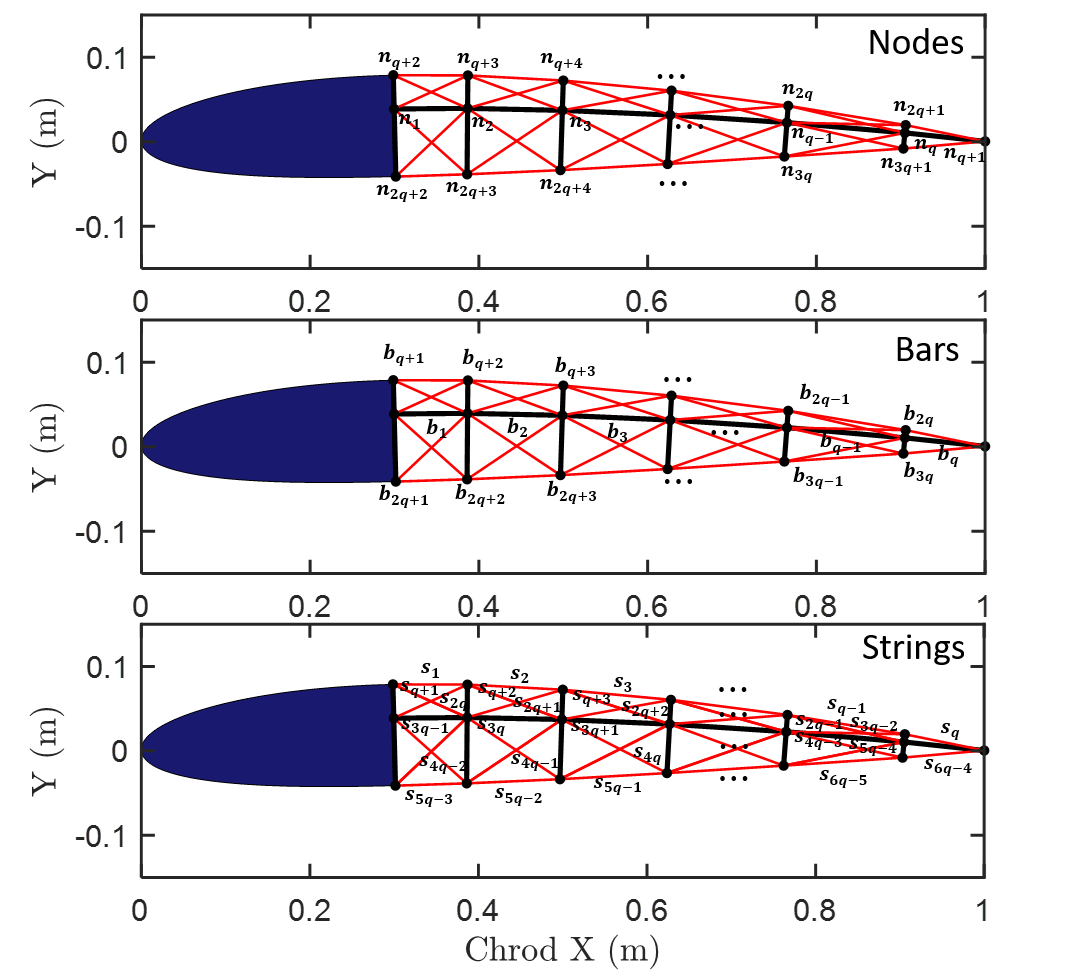}
\caption{Node, bar, and string notations of a tensegrity airfoil with complexity $q$.}
\label{topology}
\end{figure}

Inspired by the structure of vertebra, we connect the discrete points in a similar pattern, as shown in Fig. \ref{Config}. The notation of nodes, bars, and strings of a tensegrity airfoil with any complexity $q$ is given in Fig. \ref{topology}, where $q$ is the number of horizontal bars in the tensegrity structure. The discrete points on the surface of the airfoil (nodes $\bm{n}_{q+1},\bm{n}_{q+2},\cdots,\bm{n}_{3q+1}$) are determined by error bound spacing method developed in \cite{chen2020design}, which is defined as the maximum error between the continuous surface shape and each straight-line segment is less than a specified value $\delta$. The coordinate of node $\bm{n}_{i}$ ($i = 1,2,\cdots,q$) is determined by the nodes above and below this point with a same ratio $\mu \in (0,1)$, which satisfies $\bm{n}_i = \mu \bm{n}_{q+1+i} + (1-\mu)\bm{n}_{2q+1+i}$.

We define nodal, bar, string, bar connectivity, and string connectivity matrices: $N$, $B$, $S$, $C_b$, and $C_s$ to describe a tensegrity airfoil with any complexity $q$. The nodal matrix $N = [\bm{n}_1, \bm{n}_2, \cdots, \bm{n}_{3q+1}]$, its each column represents the $x$-, $y$-, and $z$-coordinate of each node ($\bm{n}_i=\begin{bmatrix}  x_i & y_i & z_i\end{bmatrix} ^T$). $C_S$ and $C_B$ are connectivity matrices (with 0, -1, and 1 contained in each column) of strings and bars. The bar and string matrices $B = [\bm{b}_1,\bm{b}_2,\cdots,\bm{b}_{3q}] = NC_b^T$, $S = [\bm{s}_1,\bm{s}_2,\cdots,\bm{s}_{6q-4}] = NC_s^T$, where $\bm{b}_j$ ($j = 1, 2, \cdots, 3q$) and $\bm{s}_k$ ($k = 1, 2, \cdots, 6q-4$) are the $j$th bar and $k$th string. $C_{b_{in}}$ and $C_{s_{in}}$ whose two elements in each row denotes the start and end node of one bar or string:
\begin{eqnarray}
\label{c_in}
C_{b_{in}}= &\begin{cases}
    [i,i+1],~1\leq i\leq q \\
   [i-q,i+1],~q+1\leq i\leq 2q \\
   [i-2q,i+1],~2q+1\leq i\leq 3q \\
   \end{cases},\\
C_{s_{in}}=& \begin{cases} 
   [i+1+q,i+2+q],~1\leq i\leq q-1 \\
   [q+i,i],~2\leq i\leq q \\
   [i,q+2+i],~1\leq i\leq q-1 \\
  [i,2q+2+i],~1\leq i\leq q-1 \\
   [2q+i,i],~2\leq i\leq q \\
  [i+1+2q,i+2+2q],~1\leq i\leq q-1 \\
      [2q+1,q+1], ~[3q+1,q+1]
   \end{cases}.
\end{eqnarray}

Then, a function $tenseg\_ind2C.m$ can be written to convert $C_{b_{in}}$ and $C_{s_{in}}$ to $C_b$ and $C_s$ \cite{Goyal2019MOTES}. The nonlinear FEM tensegrity dynamics is used as the black box system to do the system identification, which is given in a vector form \cite{ma2020fem}:
\begin{align}
\bm{M} \ddot{\bm{n}}+\bm{D} \dot{\bm{n}}+\bm{K} \bm{n}=\bm{f}_{e x}-\bm{g},\\
 \bm{M} =\frac {1} {6}(|\bm{C}|^T\hat{\bm{m}}|\bm{C}|+ \lfloor|\bm{C}|^T\hat{\bm{m}}|\bm{C}|\rfloor)\otimes \textbf{I}_3,\\
  \bm{K}=(\bm{C}^T\hat{\bm{x}}\bm{C})\otimes\textbf{I}_3,\\
  \bm{g}=\frac{g}{2}(|\bm{C}|^T\bm{m})\otimes    \begin{bmatrix}
    0 & 0 & 1
    \end{bmatrix}^T,
 \end{align}
where nodal coordinate vector $\bm{n} \in \mathbb{R}^{3 n_n}$ for the whole structure
$ \bm{n}= \begin{bmatrix} \bm{n}_1^T & \bm{n}_2^T & \cdots & \bm{n}_{3q+1}^T\end{bmatrix}^T$, connectivity matrix $\bm{C} = \begin{bmatrix}C_b^T &  C_s^T\end{bmatrix}^T$, $\bm{M}$, $\bm{D}$, and $\bm{K}$ are mass, damping, and stiffness matrices, $\bm{m}$ is the mass vector of bars and strings ($\hat{\bm{m}}$ is a diagonal matrix), $\bm{f}_{ex}$ is external forces on the structure nodes, and $\bm{g}$ is gravity vector ($g$ is gravity constant).

\section{Morphing target}\label{sec4}

\begin{figure}[htb]
\centering
\includegraphics[width=1.0\linewidth]{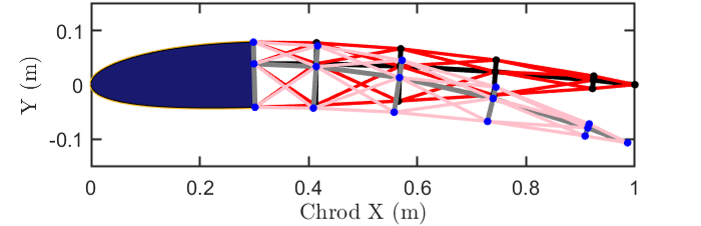}
\caption{Initial and morphing configuration of the tensegrity NACA 2412 airfoil, top one (bars in black, strings in red, and nodes in black) is the initial state, and the bottom one (bars in grey, strings in pink, and nodes in blue) is the morphing target.}
\label{ini_target}
\end{figure}

We choose NACA 2412, with a chord length $c = 1$ m, 0 $\sim$ 0.3 m as the rigid part, vertical bar length ratio $\mu = 1/3$, and error bound $\delta = 0.001$ m to generate the initial configuration of the tensegrity foil. In this example, there are five horizontal bars ($q =5$). The morphing targets are generated by the rotation of the horizontal bars (bars $\bm{b}_1, \bm{b}_2, \cdots, \bm{b}_5$) in the tensegrity structure in a linear manner while keeping the length of every bar unchanged during deformation. That is, bar $\bm{b}_1$ rotates $\theta_1 = \frac{\pi}{72}$, bar $\bm{b}_2$  rotates $\theta_2 = \frac{\pi}{36}$, and up to bar $\bm{b}_5$ rotates $\theta_5 = \frac{5\pi}{72}$, the vertical bars remain the same angle with the horizontal bars as the initial configuration. The final morphing target is shown in Fig. \ref{ini_target}.

\section{Data experiment and results}\label{sec5}

\begin{figure}[htb]
\centering
\includegraphics[width=1.01\linewidth]{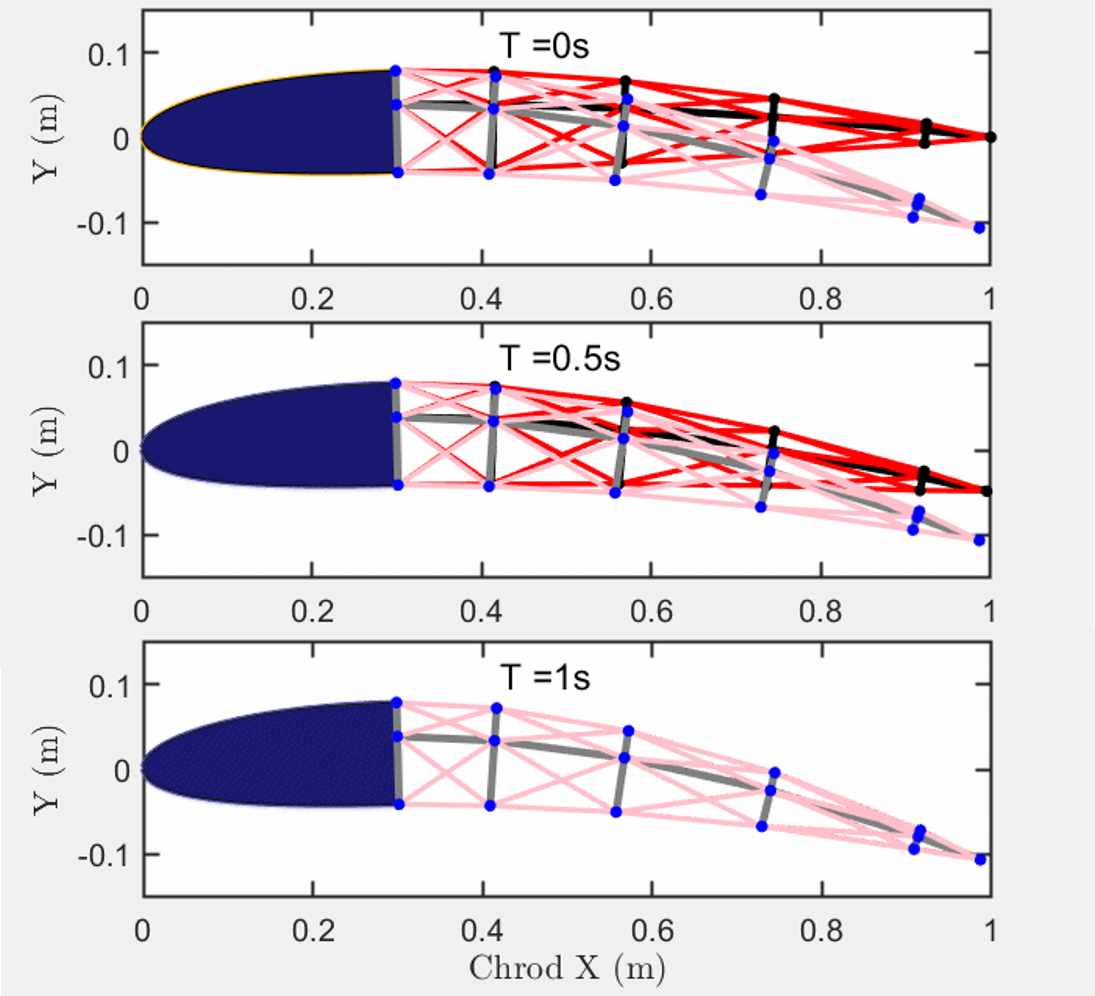}
\caption{Time history of the tensegrity morphing airfoil at T = 0s, 0.5s, and 1s.}
\label{Time_hist}
\end{figure}

\begin{figure}[htb]
\centering
\includegraphics[width=1.0\linewidth]{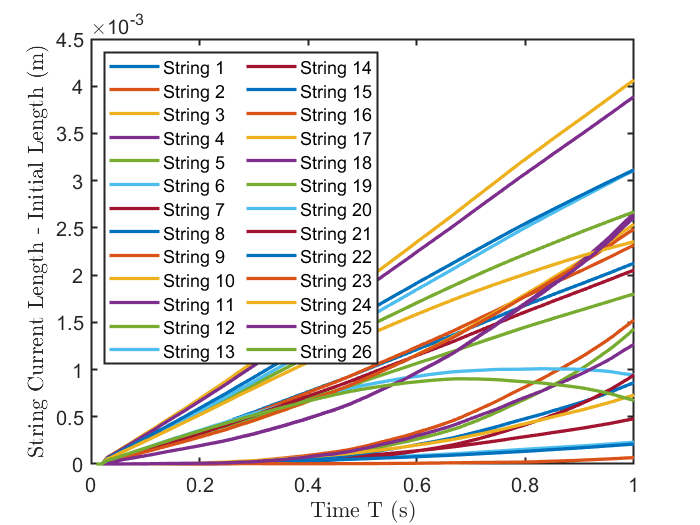}
\caption{String length time history, string current length minus string initial length v.s. time.}
\label{string_length_change}
\end{figure}

\begin{figure}[htb]
\centering
\includegraphics[width=1.0\linewidth]{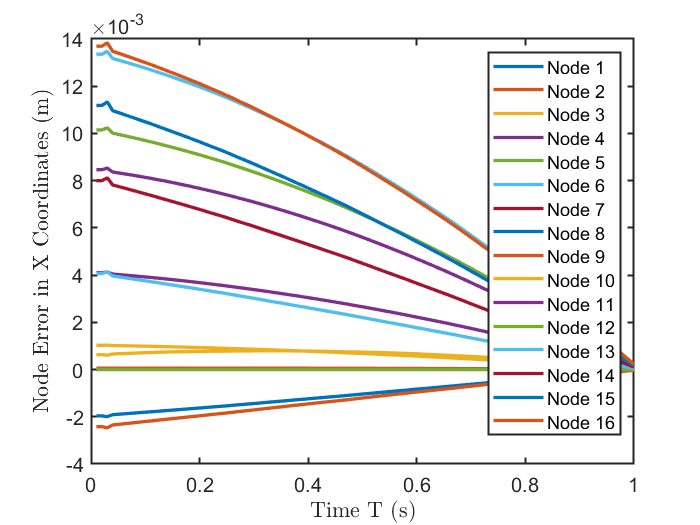}
\caption{Node error in x coordinates v.s. time.}
\label{Error_x}
\end{figure}

\begin{figure}[htb]
\centering
\includegraphics[width=1.0\linewidth]{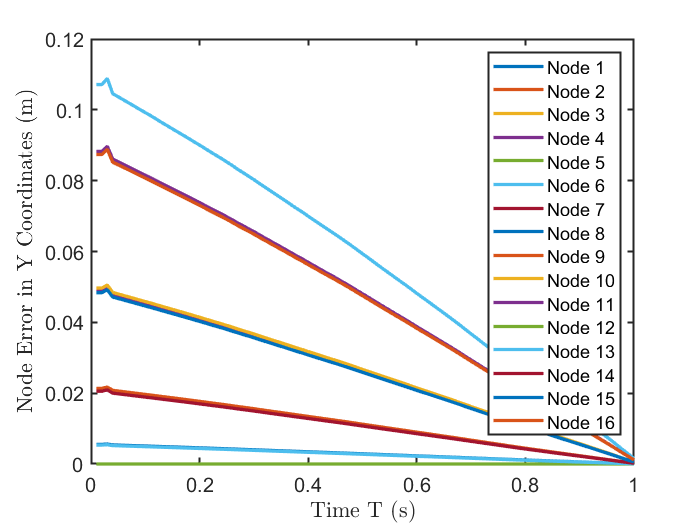}
\caption{Node error in y coordinates v.s. time.}
\label{Error_y}
\end{figure}

A black box system which contains the dynamics of the tensegrity airfoil shown in section 3 has been constructed, which accepts 26 input signal and returns 26 output signal. Its Markov parameters have been evaluated via black box experiments Eq. (\ref{H_matrix}), and control sequences have been computed using Eq. (\ref{delta_control}) to track a reference configuration. It costs 164 seconds to determine the Markov parameters from experiments, and 88 seconds for control sequence calculations using an Intel7-9700, 3.60 GHz computer.

Fig. \ref{Time_hist} shows a time history of airfoil configuration while tracking a trajectory at time T = 0s, 0.5s, and 1s. Fig. \ref{string_length_change} is the length change of each string. Fig. \ref{Error_x} and Fig. \ref{Error_y} show the error x- and y-coordinates during the process of tracking, which demonstrate the successful control of the airfoil. Note that the control starts at step 3, where one can observe an obvious bump. The first sample represents the initial conditions, and there is one sample transport delay from input to output.

% \begin{figure}[htb]
% \centering
% \includegraphics[width=1.0\linewidth]{bar_length_error.png}
% \caption{bar length error based on the nonlinear control law.}
% \label{bar_length_error}
% \end{figure}

\section{Conclusion}\label{sec6}

This letter presents a data-based optimal control design that only requires the first $N+1$ Markov parameters of a system for reference tracking applications. It optimizes the control sequence with respect to a cost function integrating tracking error and input increments. The only necessary knowledge, Markov parameters, can be evaluated in many approaches, including an input-output method. Therefore, this design does not require any explicit information about the dynamics of a system. Result demonstrates a successful control of a tensegrity morphing airfoil reference tracking application. 

% \addtolength{\textheight}{-12cm} % This command serves to balance the column lengths

%%%%%%%%%%%%%%%%%%%%%%%%%%%%%%%%%%%%%%%%%%%%%%%%%%%%%%%%%%%%%%%%%%%%%%%%%%%%%%%%

%%%%%%%%%%%%%%%%%%%%%%%%%%%%%%%%%%%%%%%%%%%%%%%%%%%%%%%%%%%%%%%%%%%%%%%%%%%%%%%%


% \section*{ACKNOWLEDGMENT}

\bibliographystyle{IEEEtran}
\bibliography{Bib}

\end{document}